\def\year{2020}\relax
\documentclass[letterpaper]{article} 
\usepackage{aaai20}  
\usepackage{times}  
\usepackage{helvet} 
\usepackage{courier}  
\usepackage[hyphens]{url}  
\usepackage{graphicx} 
\usepackage{graphicx}  
\usepackage{diagbox} 
\usepackage{epsfig}
\usepackage{amsmath}
\usepackage{amssymb}
\usepackage{color}
\usepackage[labelfont=bf]{caption}
\usepackage{multirow}
\usepackage{array}
\usepackage[dvipsnames]{xcolor}
\usepackage{caption}
\usepackage{algorithm}
\usepackage[noend]{algpseudocode}

\usepackage{subfigure}
\usepackage{verbatim}

\newcolumntype{P}[1]{>{\centering\arraybackslash}m{#1}}
\newcommand{\Paragraph}[1]{\vspace{-0mm} \noindent \textbf{#1} \hspace{0mm}}

\title{Aurora Guard: Reliable Face Anti-Spoofing via Mobile Lighting System}

\author{Jian Zhang\textsuperscript{\rm 1} 
    Ying Tai\textsuperscript{\rm 1} 
    Taiping Yao\textsuperscript{\rm 1} 
    Jia Meng\textsuperscript{\rm 1} 
    Shouhong Ding\textsuperscript{\rm 1} 
    Chengjie Wang\textsuperscript{\rm 1} \\
    \Large \textbf{Jilin Li\textsuperscript{\rm 1} 
    Feiyue Huang\textsuperscript{\rm 1} 
    Rongrong Ji\textsuperscript{\rm 2}} \\
\textsuperscript{\rm 1}Tencent YouTu Lab~~~ \textsuperscript{\rm 2}Xiamen University\\ 
\textsuperscript{\rm 1}\{timmmyzhang, yingtai, taipingyao, jeffmeng, ericshding, jasoncjwang, jerolinli, garyhuang\}@tencent.com\\
\textsuperscript{\rm 2}rrji@xmu.edu.cn\\
}

\begin{document}
	\maketitle
	
	\begin{abstract}
		Face authentication on mobile end has been widely applied in various scenarios.
		Despite the increasing reliability of cutting-edge face authentication/verification systems to variations like blinking eye and subtle facial expression, anti-spoofing against high-resolution rendering replay of paper photos or digital videos retains as an open problem.
		In this paper, we propose a simple yet effective face anti-spoofing system, termed Aurora Guard (AG).
		Our system firstly extracts the normal cues via light reflection analysis, and then adopts an end-to-end trainable multi-task Convolutional Neural Network (CNN) to accurately recover subjects' intrinsic depth and material map to assist liveness classification, along with the light CAPTCHA checking mechanism in the regression branch to further improve the system reliability.
		Experiments on public Replay-Attack and CASIA datasets demonstrate the merits of our proposed method over the state-of-the-arts.
		We also conduct extensive experiments on a large-scale dataset containing $12,000$ live and diverse spoofing samples, which further validates the generalization ability of our method in the wild.
	\end{abstract}
	
	\section{Introduction}\label{sec:introduction}
	Face anti-spoofing has been a promising topic in computer vision research, which is regarded as a very challenging problem in industry especially in remote scenarios without specific hardware equipped.
	The existing methods~\cite{yi2014face,Zhang_2019_CVPR_Workshops,Zhang_2019_CVPR} on face anti-spoofing are paying more attention on exploiting multi-modality information, $\textit{e.g.}$, RGB images, depth or infrared light.
	With the development of depth sensors, recent methods and commercial systems mainly rely on hardwares embedded with structured light (\emph{e.g.}, FaceID on iphone X), light field~\cite{xie2017one} or LIDAR to reconstruct accurate $3$D shape, which can well address the limitation of $2$D methods towards high-level security~\cite{li2016original,li2017face}. Although good anti-spoofing performance can be achieved, these methods highly rely on the customized hardware design, which unavoidably increases the system cost.
	
	\begin{figure}[tbp]
		\centering
		\setlength{\abovecaptionskip}{3mm}
		\setlength{\belowcaptionskip}{-3mm}
		\setlength{\lineskip}{\medskipamount}
		\includegraphics[width=0.98\linewidth]{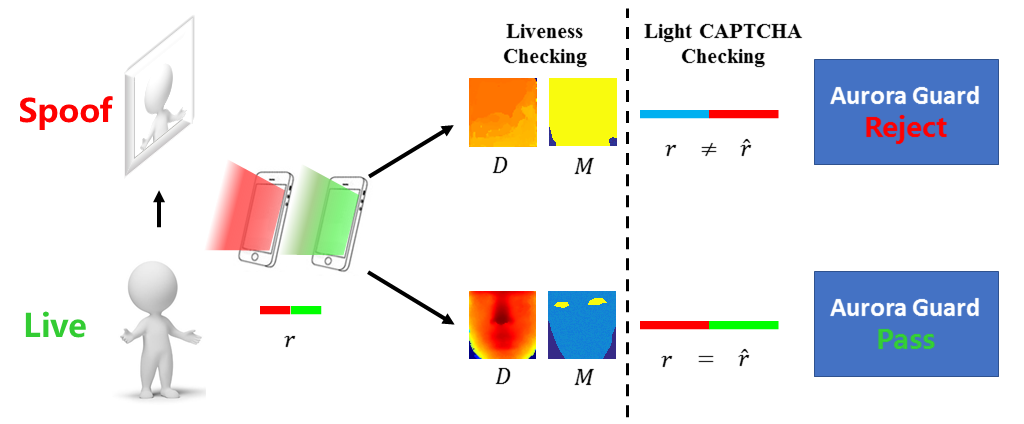}
		\caption{\small \textbf{Framework of our proposed system}.
			$D$/$M$ denotes the recovered depth/material map from the reflection frames, which improves our anti-spoofing performance against unlimited $2$D/$3$D spoofing. The whole system then imposes liveness checking on these two auxiliary information.
			$r$ denotes the light CAPTCHA generated and casted by light source and $\hat r$ denotes the light CAPTCHA estimated by our method.
			The light CAPTCHA checking mechanism further improves our system's security.}
		\label{Fig1}
	\end{figure}
	
	Considering the cost of additional sensors, recent advances on Presentation Attack Detection~(PAD) estimate depth directly from a single RGB image as a replacement.
	In particular, since $3$D reconstruction from a single image is highly under-constrained due to the lack of strong prior of object shapes, such methods introduce certain prior by recovering sparse~\cite{wang2013face} or dense~\cite{atoum2017face,Liu_2019_CVPR} depth features.
	However, on one hand, these methods still suffer from the missing of solid depth clue, leading to the lack of generalization capability.
	On the other hand, the system is easily vulnerable to $3$D attack (\emph{e.g.}, silicon/paper mask) if depth information is determinant to the final judgment.
	
	Towards solving various attacks without using additional sensors, we propose a simple, fast yet effective face anti-spoofing system termed Aurora Guard (AG).
	Its principle is using light reflection to disentangle two auxiliary information, \emph{i.e.}, depth and material, to consolidate discriminative features for real/fake classification, as shown in Fig.~\ref{Fig1}. 
	Those two information can be reliably extracted from \emph{normal cues} defined in this paper, which are the pixel-wise subtraction of two contiguous reflection frames.
	In addition, we further leverage the \emph{light CAPTCHA}, \emph{i.e.}, the random light parameters sequence, to provide an extra security mechanism by checking the consistency of our prediction with the ground truth.
	By only incorporating a single extra light source to generate the reflection frames, our method ensures both the \textit{efficiency} and \textit{portability} in a cost-free software manner, which has already been deployed on smart phones and embedded terminals that serves for \textit{millions of users}.
	
	\begin{figure*} [t!]
		\centering
		\setlength{\abovecaptionskip}{2mm}
		\setlength{\belowcaptionskip}{-1mm}
		\setlength{\lineskip}{\medskipamount}
		\includegraphics[width=0.95\linewidth]{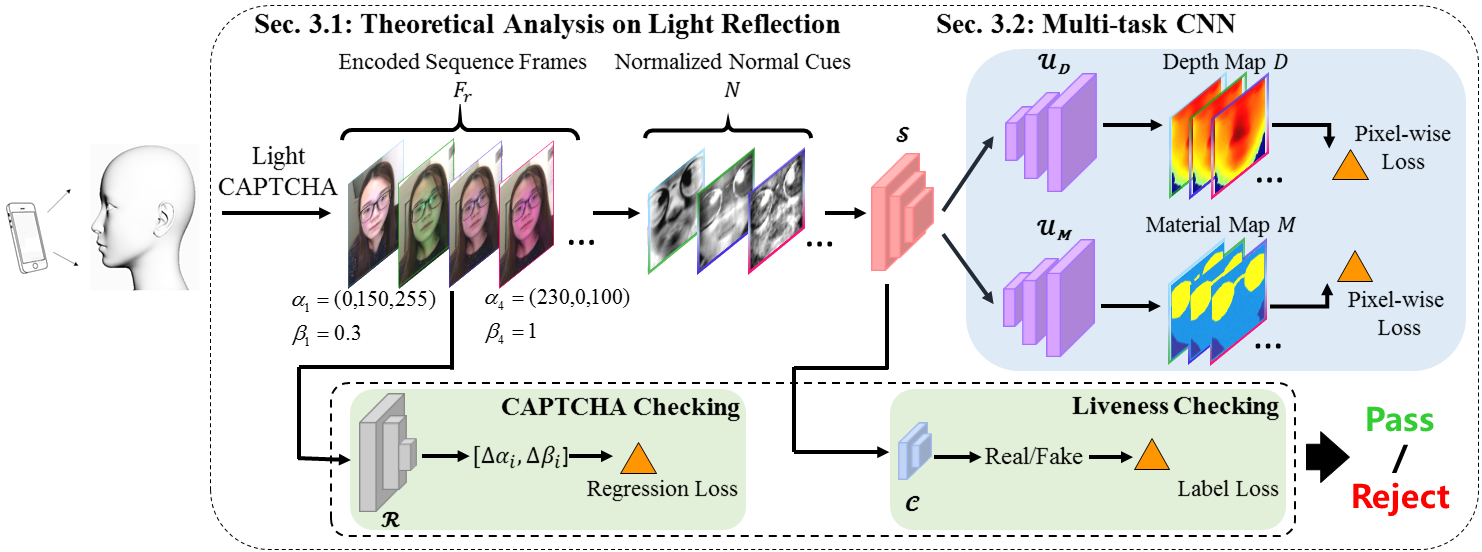}
		\caption{\small \textbf{Overview of Aurora Guard}.
			From facial reflection frames encoded by casted light CAPTCHA, we estimate the normal cues.
			In the upper-right reconstruction branches, we recover the depth maps $D$ and material maps $M$ from the encoded featrues of normal cues via two separate decoders $\mathcal{U_D}$ and $\mathcal{U_M}$.
			In the classification net $\mathcal{C}$, we utilize the consolidated encoded features to perform liveness checking. In the regression net $\mathcal{R}$, we obtain the estimated light CAPTCHA for double checking.}
		\label{Pipeline}
	\end{figure*}
	
	In particular, our method consists of three parts:
	($1$) 
	We adopt the Lambertian model to cast dynamic changing light specified by the random light CAPTCHA, and then extract the normal cues from every two contiguous reflection frames. The solid depth and material information are then embodied in the normal cues.
    ($2$) We use a compact encoder-decoder structure to conduct disentanglement of depth and material simultaneously. 
	With two regression branches recovering depth and material maps respectively, the learned features are robust for both $2$D and $3$D attacks, which facilitate the liveness judgment in the classification branch.
	($3$) We provide an additional branch to estimate the light parameter sequence, which forms a light CAPTCHA checking mechanism to handle the special attack named \textit{modality spoofing}, a very common attack in real scenarios.
	
	Moreover,
	since the imaging qualities (resolution, device) and the types of Presentation Attack Instruments (PAI) are essential to the performance evaluation of practical face authentication, we further build a dataset containing videos of facial reflection frames collected by our system, which is the \textit{most comprehensive} and \textit{largest} one of its kind compared with other public datasets.
	On this dataset, we demonstrate that our depth reconstruction is competitive to the professional $3$D sensor qualitatively and quantitatively. Also, our material reconstruction serves as a powerful tools to block a large proportion of $3$D attacks. 
	As a result, without extra hardware designs, our model achieves comparable performance against the expensive hardware on face anti-spoofing.
	
	To sum up, the main contributions of this work include:
	
	$\bullet$ A simple, fast yet effective face anti-spoofing method is proposed, which is practical in real scenarios \textit{without} the requirement on specific depth hardwares. 
	
	$\bullet$ A \textit{cost-free} disentangle net is proposed to recover the depth and material maps via the \textit{normal cues} extracted from two contiguous reflection frames for liveness classification.
	
	$\bullet$ A novel \textit{light CAPTCHA checking mechanism} is proposed to significantly improve the security against the attacks, especially the modality spoofing.
	
	$\bullet$ A dataset containing comprehensive spoof attacks on various imaging qualities and mobile ends is built.

	\section{Related Work}\label{sec:relatedwork}
	We review the related work from four perspectives, and summarize the key difference of our method in {Tab.~\ref{tab:related}}.
	
	\Paragraph{Local Texture based Methods.}
	The majority of common presentation attacks are the recaptured images shown on printed photo and screens, in which the textures are different from the original ones and can be leveraged for face anti-spoofing.
	For example, 
	\cite{DBLP:journals/tifs/WenHJ15} adopted image distortion information as countermeasure against spoofing.
	\cite{li2017face} proposed Deep Local Binary Pattern (LBP) to extract LBP descriptors on convolutional feature map.
	\cite{boulkenafet2017face} converted the face image from RGB space to HSV-YCbCr space and extracted channel-wise SURF features~\cite{bay2006surf} to classify liveness result.
	However, since the above methods operate on $2$D images, they suffer from poor generalization to unseen attacks and complex lighting conditions, especially when RGB sensors have low resolution or quality.
	In contrast, our method exploits material information (\emph{e.g.}, the intrinsic albedo) via the reflection increments from RGB images, which is more robust and more accurate to various attacks.

	\Paragraph{Depth Sensor based Methods.}
	It is well known that the $3$D facial cues can be used to defeat $2$D presentation attacks.
	For example,~\cite{wang2017robust} directly exploited depth sensors such as Kinect to obtain depth map, which is combined with texture features to conduct anti-spoofing.
	\cite{xie2017one} introduced a light field camera to extract depth information from multiple refocused images took in one snapshot.
	Moreover, iPhone X incorporates a structured-light sensor to recover accurate facial depth map, which obtains impressive performance.
	However, although iPhone X achieves high accuracy, there are two practical problems.
	First, it uses an \textit{expensive} $3$D camera to obtain accurate depth.
	Second, its implementation details are missing.
	In contrast, our method has competitive results against $3$D hardware via a \textit{cost-free} depth recover net, and is easy to follow for \textit{re-implementation}.
	
	\begin{table} [t!]
		\small
		\centering
		\setlength{\abovecaptionskip}{1mm}
		\vspace{1.5mm}
		\scalebox{0.71}{
			\begin{tabular}{|c|c|c|c|c|c|}
				\hline
				Method   & Local Texture    & Depth Sensor & Depth from Single Image & Ours \\ \hline
				Depth    & $\pmb \times$    & $\pmb \surd$ & $\pmb \surd$            & $\color{red}{\pmb \surd}$ \\ \hline
				High accuracy  & $\pmb \times$   & $\pmb \surd$  & $\pmb \times$     & $\color{red}{\pmb \surd}$\\ \hline
				Hardware-Free  & $\pmb \surd$    & $\pmb \times$ & $\pmb \surd$      & $\color{red}{\pmb \surd}$\\ \hline
				Real-Time on CPU  & $\pmb \surd$  & $\pmb \times$  & $\pmb \surd$    & $\color{red}{\pmb \surd}$\\ \hline
			\end{tabular}
		}
		\caption{\small Comparisons with related methods.}
		\label{tab:related}
		\vspace{-2.5mm}
	\end{table}
	
	\Paragraph{Depth Estimated from Single Image.}
	\cite{wang2013face} firstly attempted to recover a sparse $3$D facial structure from RGB image for face anti-spoofing.
	\cite{atoum2017face} proposed a two-stream depth-based CNN to estimate both texture and depth.
	Recently,~\cite{liu2018learning} fused multiple sequential depth predictions to regress to a temporal rPPG signal for liveness classification.
	However, $3$D reconstruction from a single image is still highly under-constrained, since these methods suffer from missing solid $3$D information clue.
	As a result, their anti-spoofing classifiers are hard to generalize to unseen attacks, and are also sensitive to the quality of RGB camera.
	To address the inaccurate depth issue, our method first obtains normal cues based on light reflection, which better removes the effects of illuminance.
	Then a compact encoder-decoder network is trained to accurately recover the depth map. 
	
	\Paragraph{Lambertian Reflection based Methods.}
	\cite{tan2010face} first identified the importance of Lambertian modeling for face anti-spoofing, and obtained rough approximations of illuminance and reflectance parts.
	\cite{chan2018face} adopted Lambertian reflection model to extract simple statistics (\emph{i.e.}, standard deviation and mean) as features, and achieved further performance gain.
	Our method differs from the above methods in three aspects:
	($1$) We \textit{actively} perform light reflection via an extra light source specified by random light parameter sequence, while the above methods do NOT.
	($2$) We introduce a novel \textit{light CAPTCHA} checking mechanism to make the system more robust, while the above methods lack such scheme again.
	($3$) We incorporate deep networks to learn powerful features, while the above methods use simple handcrafted features.
	
	\section{The Proposed Method}\label{sec:proposed_method}
	Fig.~\ref{Pipeline} illustrates the flow chart of the proposed method.
	Specifically, we first set a smart phone (or any other devices) with front camera and light source ($\textit{e.g.}$, the screen) in front of the subject.
	Then, a random parameter sequence ($\textit{i.e.}$, light CAPTCHA) of light hues and intensities is generated, \emph{i.e.}, $r=\{(\alpha_i, \beta_i)\}_{i=1}^{n}$, with $n$ frames.
	We manipulate the screen to cast dynamic light specified by the light CAPTCHA $r$.
	After the reflection frames $F_r$ are captured, we sequentially extract the normal cues from every two contiguous frames,
	which are the inputs of a multi-task CNN to predict liveness label and regress the estimated light CAPTCHA $\hat r$. The final judgment is been made from both of the predicted label and the matching result between $\hat r$ and $r$.
	
	\subsection{Theoretical Analysis on Light Reflection}
	\label{3.1}
	
	Since objects with rough surfaces (\emph{e.g.} human face) are diffuse reflectors, light casted onto surface point is scattered and reflected, and then perceived as the final imaging in the camera.
	Given images containing reflection on the object surface, we measure the magnitude variations among different images, under the assumption of Lambertian reflection model\footnote{Our system works well when LUX (SI derived unit of illuminance) is under $800$, which can be satisfied in many real scenarios.} with a weak perspective camera projection.
	
	In particular, Lambert's Law regards the reflected part to be equal on all directions on the diffuse surface.
	In other words, for any pixel point $\mathbf{p}$ of the camera image under specific casting light $L_{r}$, its intensity $F_{r}(\mathbf{p})$ is formulated as:
	
	\begin{equation}
		F_{r}(\mathbf{p}) = \rho_{p}(k_{a}+k_{r}\mathbf{l}\cdot \mathbf{n}_{p}),
	\end{equation}
	
	where $k_{a}$ is the ambient weight, $k_{r}$ is the diffuse weight, $\mathbf{l}$ is the light source direction, $\rho_{p}$ is the albedo and $\mathbf{n}_{p}$ is the point normal.
	When light changes {suddenly}, $k_{a}$ and $\mathbf{l}$ (position of the screen) are not supposed to change temporally and can be regarded as constants. We adopt affine transformation to align $\mathbf{p'}$ and $\mathbf{p}$ between image pairs, with transformation matrix estimated from the facial landmarks detected by PRNet~\cite{feng2018prn}. Then in another image under casting light $L_{r'}$, the intensity of the registered pixel $\mathbf{p'}$ is:
	
	\begin{equation}
		F_{r'}(\mathbf{p}) = F_{r'}(\mathbf{p'}) = \rho_{p'}(k_{a}+k_{r'}\mathbf{l}\cdot \mathbf{n}_{p'}).
	\end{equation}
	By calculating pixel-wise subtraction of these two images, we attain the scalar product $N_{\Delta r}(\mathbf{p})$ on each point:
	
	\begin{equation}
		N_{\Delta r}(\mathbf{p}) = \frac{F_{r}(\mathbf{p}) - F_{r'}(\mathbf{p})}{k_{r} - k_{r'}} = \rho_{p} \mathbf{l} \cdot \mathbf{n}_{p} = \rho_{p} \cdot \cos\theta_p,
	\end{equation}
	where the scalar map arranged by $N_{\Delta r}(\mathbf{p})$ is the \textit{normal cue}, and $\theta_p$ indicates the angle between the light source direction and the point normal.
	
	Comparing a single reflection frame with the normal cue, we address the following two issues: 1) One potential weakness of a single frame is its sensitivity to environment changes. 
	The normal cue is thus an alternative, since the environment bias imposes the same intensity on two contiguous frames and is then cancelled out by a pixel-wise subtraction. 2) The normal cue contains and only contains two representations (\emph{i.e.}, the albedo $\rho_{p}$ and the angle $\theta_p$), which are strong indicators of objects' material and depth, respectively.
	In light of the Lambertian model, we are confirmed that the normal cue is superior to a single frame for disentangling these two solid cues. 
	
	\subsection{Multi-task CNN}
	
	After obtaining $m$ normal cues $N_1, N_2,...,N_m$ of one video, we adopt a multi-task CNN that has three submodules to achieve depth/material reconstruction, liveness classification and light CAPTCHA regression, respectively.
	Note that our multi-task structure is \textit{task-driven}, which enables double checking mechanism to improve the robustness on modality spoofing in practical scenarios.
	
	\Paragraph{Depth/Material Reconstruction.}
	As analyzed above, the normal cues extracted from facial reflection frames contain two kinds of semantic information: depth and material.
	To efficiently split these two kinds of features, we adopt a shared encoder network with two separated decoder branches to recover depth and material maps, respectively.
	In order to balance the performance and speed, the encoder is cut from ResNet-$18$~\cite{he2016deep} to finish a $32\times$ downsampling, while the decoder is inspired by~\cite{ronneberger2015u,tai2018fsrnet}, in which we use the inverted residual block~\cite{sandler2018mobilenetv2} to conduct precise upsampling. 
	The recovered maps are then sent for a pixel-wise supervision, which guide the network to disentangle solid depth and material maps from the normal cues.
	
	After obtaining $m$ estimated depth maps: $D_1, D_2, ..., D_m$ and material maps: $M_1, M_2, ..., M_m$ of the video, the reconstruction net has the following loss function: 
	
	\vspace{-2mm}
	\begin{equation}\
		\small
		\begin{aligned}
			\mathcal{L}_{rec} = & \frac{1}{m} \sum_{i=1}^{m} \Big \{ 
			\lambda_{dep} 
			\sum_{\mathbf{p} \in \mathbb{Z}^2} - log
			(
			e^{d_k(\mathbf{p})} /
			(
			\begin{matrix}
				\sum_{k'=1}^{16} e^{d_{k'}(\mathbf{p})}
			\end{matrix}
			)
			) \\
			& + \lambda_{mat}
			\sum_{\mathbf{p} \in \mathbb{Z}^2} - log
			(
			e^{d_l(\mathbf{p})} /
			(
			\begin{matrix}
				\sum_{l'=1}^{4} e^{d_{l'}(\mathbf{p})}
			\end{matrix}
			)
			)
			\Big \},
		\end{aligned}
		\label{eq:ReconLoss}
	\end{equation}
	where $k: \Omega \to {1, ..., 16}$ and $l: \Omega \to {1, ..., 4}$ are the ground truth depth and material labels, $d_k(\mathbf{p})$ and $d_l(\mathbf{p})$ are the corresponding feature map activation on channel $k$ or $l$ at the pixel position $\mathbf{p}$. 
	In both recovering branches, we adopt $2$D pixel-wise softmax over the predicted map combined with the cross-entropy loss function.
	
\begin{figure}[t!]
		\centering
		\setlength{\abovecaptionskip}{2mm}
		\setlength{\belowcaptionskip}{-3mm}
		\setlength{\lineskip}{\medskipamount}
		\includegraphics[width=0.98\linewidth]{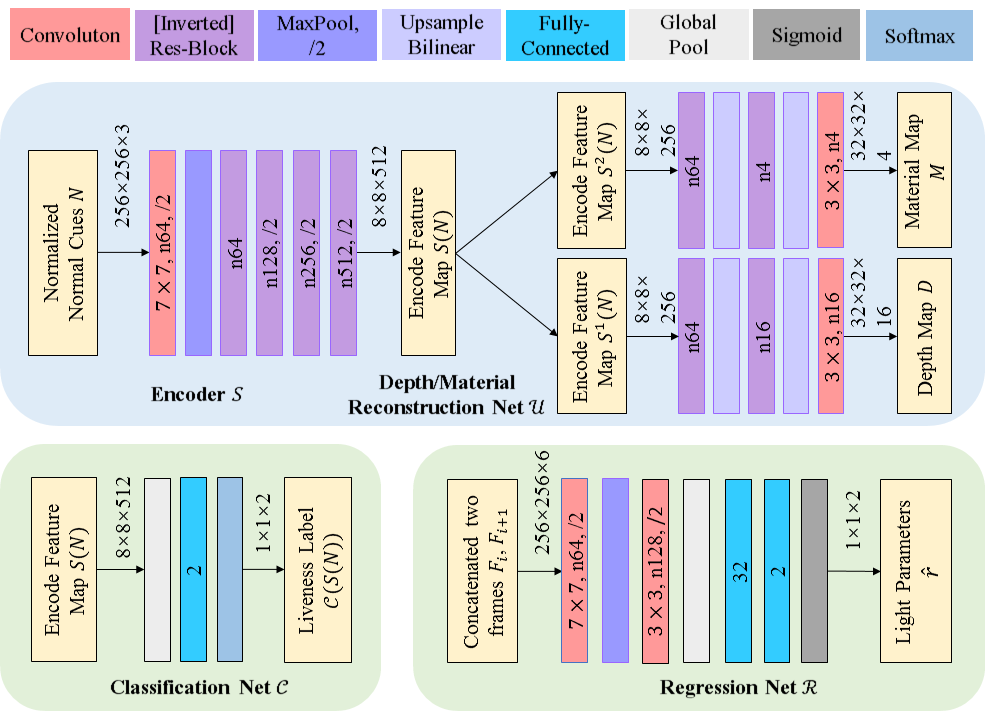}
		\caption{\small \textbf{The architecture details of the proposed multi-task CNN}. Here $n$ denotes the number of output feature maps.}
		\label{DepthMapNet}
	\end{figure}

	\Paragraph{Liveness Classification.}
	The reconstruction submodule consolidates extracted features from the encoder, which facilitates further authentication by explicitly splitting depth and material information.
	Since depth is robust enough to identify $2$D attacks and material serves as an auxiliary tool to locate $3$D attacks,  
	the extracted feature map can distinguish the real face from various presentation attacks only via a simple classification structure.
	Detailed structures of reconstruction and classification submodule are shown in Fig.~\ref{DepthMapNet}.
	The classifier has the following loss function:
	
	\vspace{-2mm}
	\begin{equation}\
		\small
		\begin{aligned}
			\mathcal{L}_{cls} = & \frac{1}{m} \sum_{i=1}^{m} \Big \{ -c_ilog(\mathcal{C}(\mathcal{S}(N_i))) \\
			& - (1-c_i)log(1-\mathcal{C}(\mathcal{S}(N_i)))
			\Big \},
		\end{aligned}
		\label{eq:ClsDepthLoss}
	\end{equation}
	where $\mathcal{S}$ denotes the stem operation that contains a clipped ResNet-18, $\mathcal{C}$ denotes the liveness prediction net, and $c_i$ is the liveness label of the $i$-th normal cue.
	
	\Paragraph{Light Parameter Regression.}
	\label{Regression}
	Moreover, we reinforce our system's security against \textit{modality spoofing}
	by customizing the casted light CAPTCHA, and exploit a regression network to decode it back for automatical double checking.
	
	By feeding two contiguous reflection frames as the input, the regression net has the loss function $L_{reg}$ as:
	
	\begin{equation}\
		\small
		\begin{aligned}
			\mathcal{L}_{reg} = & \frac{1}{m} \sum_{i=1}^{m} \{ \|\mathcal{R}(F_i, F_{i + 1}) - \Delta r_i\|^{2} \},
		\end{aligned}
	\end{equation}
	where $\mathcal{R}$ denotes the regression net, $\Delta r_i$ is the ground truth light parameter residual of reflection frames $F_{r_{i}}$ and $F_{r_{i + 1}}$.
	
	Suppose there are $V$ videos in the training set, the entire loss function of our multi-task CNN is formulated as:
	
	\vspace{-3mm}
	\begin{equation}\
		\small
		\begin{aligned}
			\mathcal{L}{(\Theta)} = & \mathop{\arg\min}\limits_{\mathbf{\Theta}} \frac{1}{2V} \sum_{v=1}^{V} \{ \mathcal{L}_{rec}^v + \lambda_{cls}\mathcal{L}_{cls}^v + \lambda_{reg}\mathcal{L}_{reg}^v \},
		\end{aligned}
	\end{equation}
	where $\Theta$ denotes the parameter set, $\lambda_{cls}$ is the weight of classification loss, $\lambda_{reg}$ is the weight of CAPTCHA regression loss.
	In practice, we set the light CAPTCHA to be composed by $4$ types of lights in random order, which balances the robustness of CAPTCHA checking and time complexity. 
	
	We show the overall decision pipeline in Alg.~\ref{alg:AG}, which is a video-wise procedure:
	We set the rate of light changing identical to the frame rate, thus the frames hold different light reflections.
	The length of $F, r$ equals to $m+1$.
	For the classification net, the output softmax scores are compared with the predefined $\tau_{cls}$. A consensus is obtained if and only if at least half of the scores exceed the threshold.  
	For the regression net, the Signal-to-Noise Ratio ($SNR$) is adopted to check if the estimated light parameter sequence matches the ground truth sequence (\emph{i.e.}, $SNR$ is larger than $\tau_{reg}$).
	During the test phase, the final judgement is mutually determined by both branches.
	
	\begin{algorithm}[t!]
		\small
		\caption{The Video-wise Pipeline}
		\label{alg:AG}
		\begin{algorithmic}[1]
			\Procedure{AntiSpoofing}{$F, \hat{r}, \hat{D}, \hat{M}, \hat{c}, train$}
			
			\State $cnt \gets 0$
			
			\For{$i = 1 \to m$}
			\State $F_{i} \gets$ \Call{WarpAlign}{$F_{i}, F_{i + 1}$}
			\State $N_{i}\gets \frac{F_{i} - F_{i + 1}}{k_{\hat{r}_{i}} - k_{\hat{r}_{i + 1}}}$
			\State $S_{i} \gets \mathcal{S}(N_{i})$ \Comment{Shared encoder}
			\State $S_{i}^1, S_{i}^2 \gets$ \Call{Bisect}{$S_{i}$}
			\State $D_{i} \gets \mathcal{U_D}(S_{i}^1)$ \Comment{Recovered depth map}
			\State $M_{i} \gets \mathcal{U_M}(S_{i}^2)$ \Comment{Recovered material map}
			\State $c_{i} \gets \mathcal{C}(S_{i})$ \Comment{Classification score}
			\If {$train$}
			\State update $\mathcal{L}_{rec}$ from $D_{i}, \hat{D}_{i}$ and $M_{i}, \hat{M}_{i}$
			\State update $\mathcal{L}_{cls}$ from $c_{i}, \hat{c}_{i}$
			\EndIf
			
			\State $\Delta r_{i} \gets \mathcal{R}(F_{i}, F_{i + 1})$ \Comment{Estimated light parameter}
			\If {$train$}
			\State update $\mathcal{L}_{reg}$ from $\Delta r_{i}, (\hat{r}_{i + 1} - \hat{r}_{i})$
			\EndIf
			
			\If {$c_{i} > \tau_{cls}$  }  
			\State $cnt \gets cnt+1$
			\EndIf
			
			\EndFor
			
			\If {\textbf{not} $train$}
			\State $SNR \gets$ \Call{CalcSNR}{$r, \hat{r}$}
			\If {$cnt >  \frac{m}{2}$ \textbf{and} $SNR > \tau_{reg}$}
			\State \textbf{return} $live$
			\Else
			\State \textbf{return} $spoof$
			\EndIf
			\EndIf
			\EndProcedure
		\end{algorithmic}
	\end{algorithm}
	
	\subsection{Dataset Collection}
	\label{Dataset}

\begin{figure}[t!]
		\centering
		\setlength{\lineskip}{\medskipamount}
		\setlength{\abovecaptionskip}{1mm}
		\includegraphics[width=0.98\linewidth]{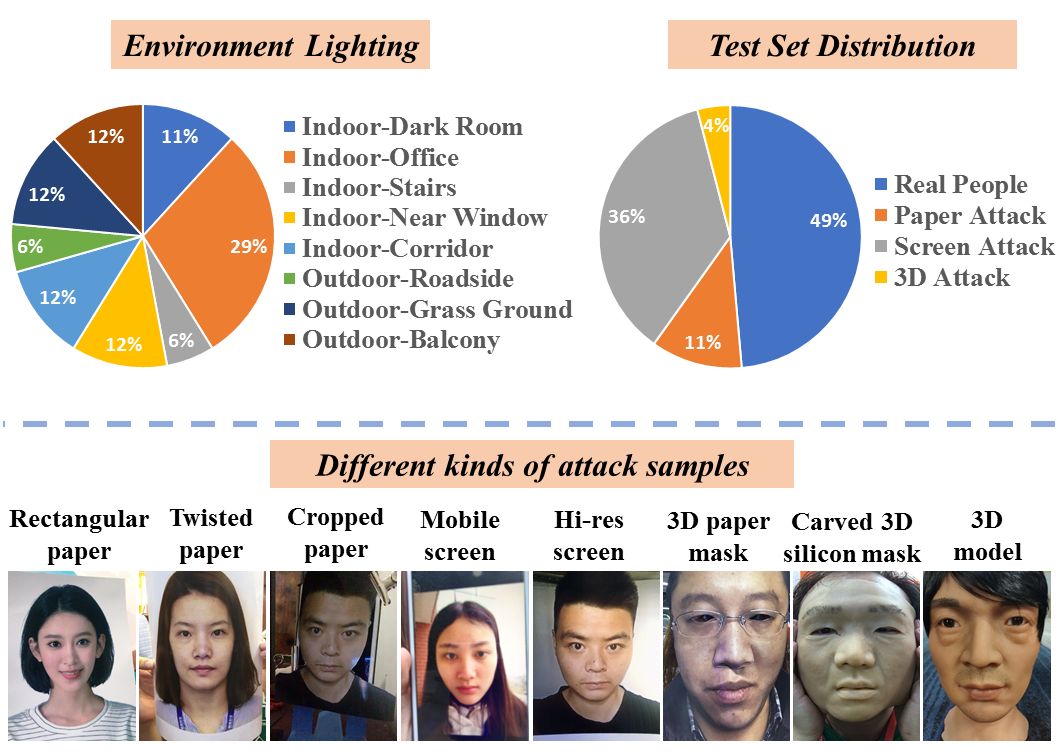}
		\caption{\small \textbf{Statistics and attack samples} of our dataset.}
		\label{AttackSamples}
	\end{figure}

	Various imaging qualities and the types of PAIs are very important for practical remote face authentication.
	To address this need, we collect a new dataset, in which each data sample is obtained by casting dynamic light sequence onto the subject, and then record the $30$-fps videos.
	Some statistics of the subjects are shown in Fig.~\ref{AttackSamples}.  
	Note that we mainly collect $2$D attacks, which are the main target in most prior anti-spoofing methods~\cite{atoum2017face,liu2018learning} as the cost to produce and conduct $3$D attacks in real scenarios is much higher than $2$D attacks.
	Besides, $3$D attacks are still essential components in our dataset, to support the situation where depth information is not sufficient for final judgment.
	
	Compared to the previous datasets~\cite{casia2012face,DBLP:conf/biosig/ChingovskaAM12,liu2018learning}, our dataset has three advantages:
	($1$) It is the \textit{largest} one that includes $12,000$ live and spoof videos, with average duration to be $3$s, collected from $200$ subjects. In contrast, the dataset in~\cite{liu2018learning} has $4,620$ videos from $165$ subjects.
	($2$) It uses the \textit{most types} of PAIs (\textit{i.e.}, $50$ smart phones, compared to $4$ in~\cite{liu2018learning}) to obtain good simulation of real-world mobile verification scenarios.
	($3$) It contains the \textit{most comprehensive} attacks that include various print, replay, modality and another spoof face by light projector (see Fig.~\ref{AttackSamples}). 
	
	We divide samples into $3$ parts through the spoof types: paper attack, screen attack and other complex 3D attacks consisting of carved paper masks, carved silicon masks, \emph{etc}.
	In each part, the data is split proportionally into train set, validation set and test set.
	Moreover, the amounts of live data and spoof data stay equal in our dataset.
	The live data is collected under multiple variations including interference illumination on face, noisy imaging quality and different poses. 
	The spoof data are collected through abundant PAIs.

	\section{Experiments} \label{sec:exp}
	\subsection{Implementation Details}
	\Paragraph{Model Training.}
	We use Pytorch to implement our method and initialize all convolutional and fully-connected layers with normal weight distribution~\cite{kaimingnorm}.
	For the optimization solver, we adopt RMSprop~\cite{RMSprop} in training.
	Training our network roughly takes $5$ hours using a single NVIDIA Tesla P$100$ GPU and iterates for $\thicksim$$300$ epochs.
	
	\Paragraph{Evaluation Criteria.}
	We use common criteria to evaluate the anti-spoofing performance, including False Rejection Rate ($FRR$), False Acceptance Rate ($FAR$) and Half Total Error Rate ($HTER$), which depends on the threshold value $\tau_{cls}$.
	To be specific, $FRR$ and $FAR$ are monotonic increasing and decreasing functions of $\tau_{cls}$, respectively.
	A more strict classification criterion corresponds to a larger threshold of $\tau_{cls}$, which means spoof faces are less likely to be misclassified.
	For certain data set $\mathbb{T}$ and $\tau_{cls}$, $HTER$ is defined as:
	
	\vspace{-1.5mm}
	\begin{equation}
		\small
		HTER(\tau_{cls},\mathbb{T})=\frac{FRR(\tau_{cls},\mathbb{T})+FAR(\tau_{cls},\mathbb{T})}{2}\in(0,1).
	\end{equation}
	For our reported $HTER$ on test set, the value of $\tau_{cls}$ is determined on the Equal Error Rate ($EER$) using the validation set, where the $EER$ is the $HTER$ subjected to that $FAR$ equals $FRR$.
	
	\begin{figure}[t!]
		\centering
		\setlength{\lineskip}{\medskipamount}
		\setlength{\abovecaptionskip}{1mm}
		\subfigure[real vs screen] {
			\includegraphics[width=0.3 \linewidth]{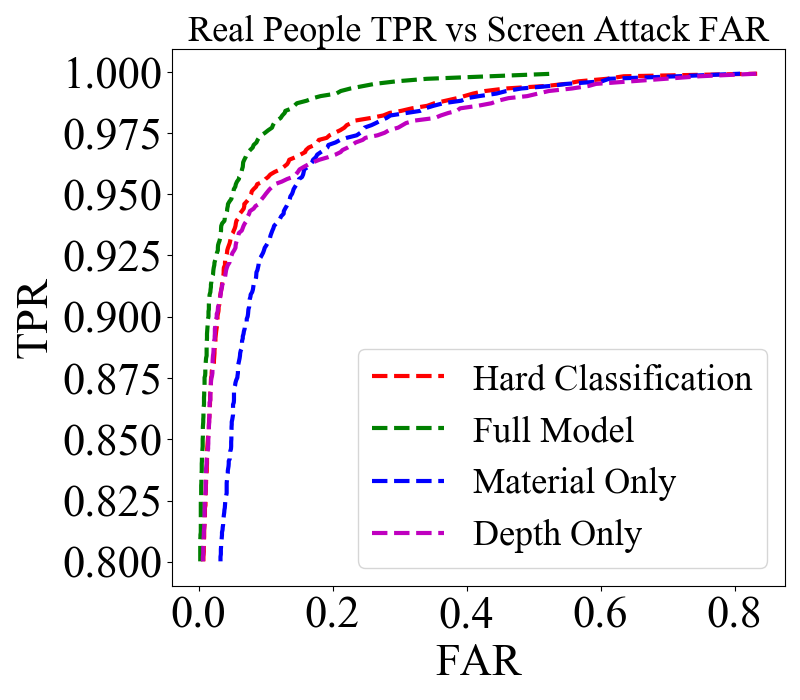}
		}
		\subfigure[real vs paper] {
			\includegraphics[width=0.3 \linewidth]{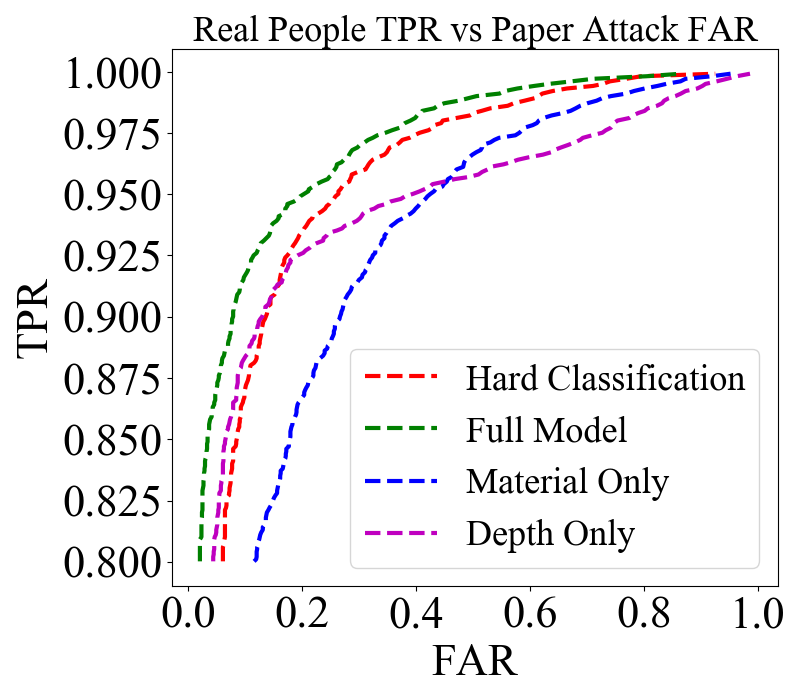}
		}
		\subfigure[real vs 3D] {
			\includegraphics[width=0.3 \linewidth]{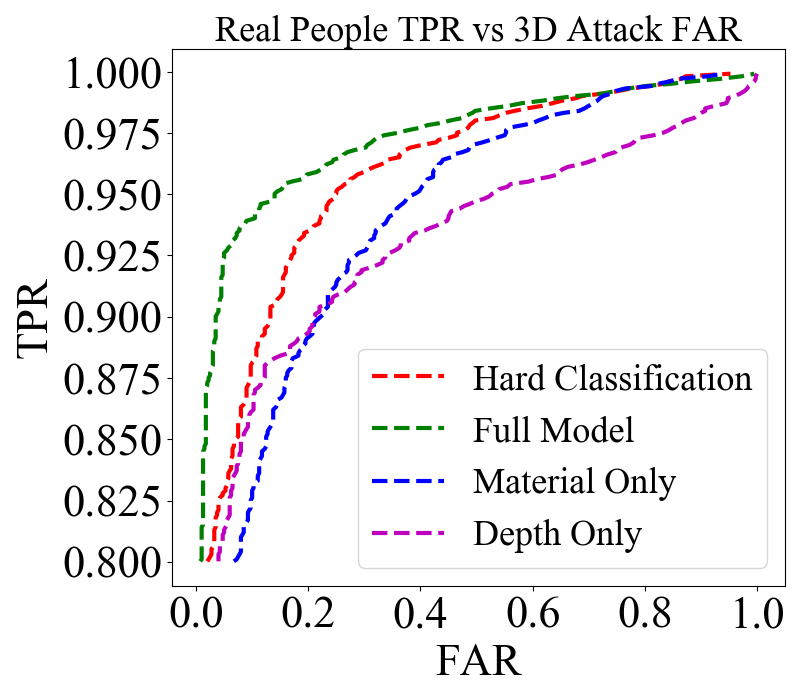}
		}
		\caption{\small \textbf{Verification on depth/material disentanglement.} Each figure shows 4 pipelines performance under distinct attacks.}
		\label{ablation_1}
		\vspace{-1mm}
	\end{figure}
	
	\subsection{Ablation Study}
	\Paragraph{Effectiveness of Depth/Material Disentanglement.\iffalse Supervision \fi}
	First, we evaluate the significance of disentanglement in depth and material information. To be specific, we construct another three pipelines for comparison, which utilize neither or either depth and material supervision, to discriminate real people from certain attacks. By adjusting the threshold, we report the $ROC$ curves under four settings,
	as shown in Fig.~\ref{ablation_1}.
	Note that despite the extracted normal cues support a strong baseline for hard classification, our full model with disentanglement still surpasses the original one. However, if we use either depth or material information only as supervision, the corresponding model fails in generalizing to the test set, which demonstrates that depth and material are correlated and extracting only one of them incurs overfitting. 
	From the comparison of $EER$ rate of bottom two curves, there is also a strong evidence for the hypothesis that depth information is useful for blocking $2$D attacks while the material information is well-performed for recognizing $3$D attacks.
	
\begin{figure}[t!]
		\centering
		\setlength{\lineskip}{\medskipamount}
		\setlength{\abovecaptionskip}{2mm}
		\setlength{\belowcaptionskip}{0mm}
		\includegraphics[width=0.98\linewidth]{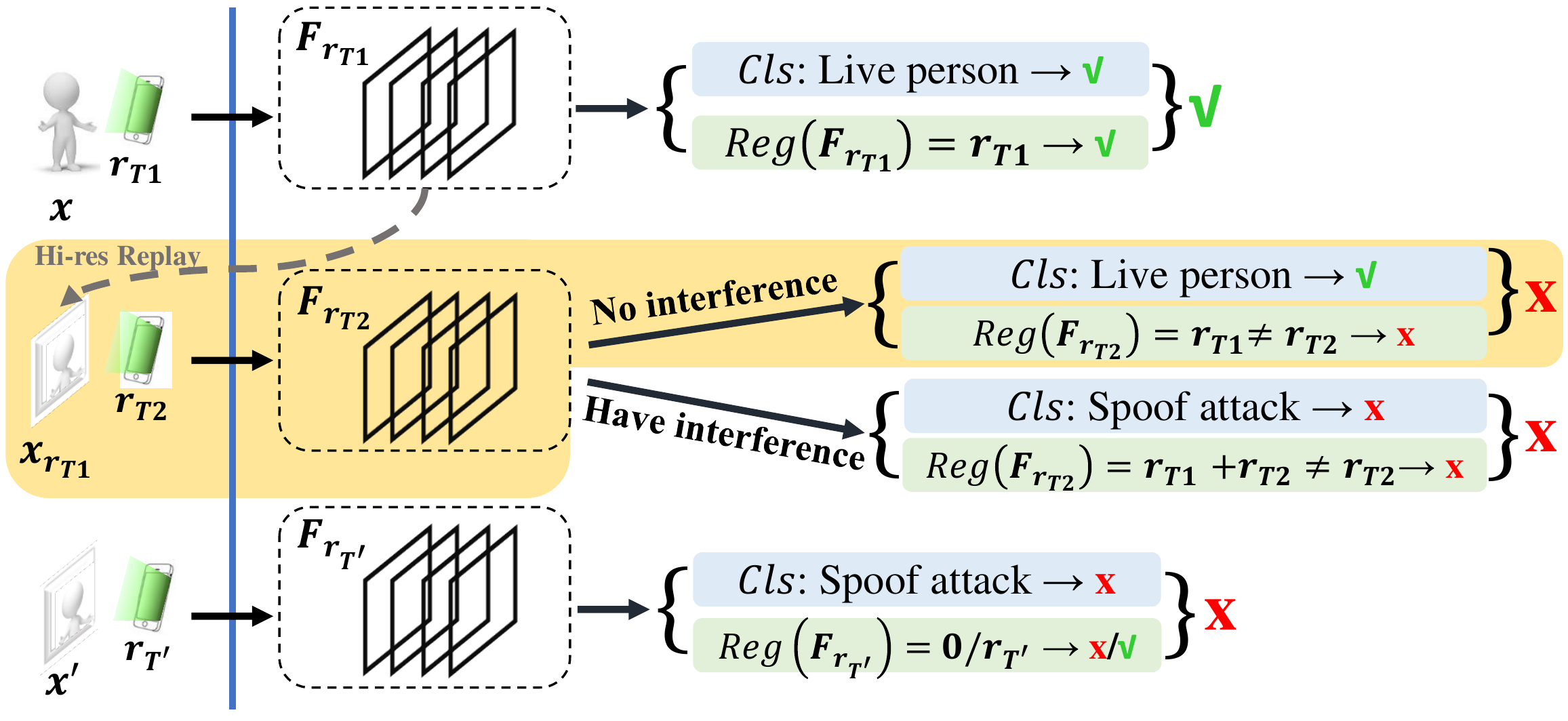}
		\caption{\small \textbf{Illustration on our double checking mechanism}.
			$Cls, Reg$ are the classification net and regression net, respectively.
			$1$) The first row handles live person.
			$2$) The highlighted yellow part in the second row represents \textit{modality spoofing} (\textit{i.e.}, $x_{r_{T1}}$), which replays the formerly captured Hi-res video frames $F_{r_{T1}}$ that contains true facial reflection, which fools the $Cls$ but can be defended by the light CAPTCHA checking scheme in $Reg$.
			$3$) No interference indicates the reflection effect caused by $r_{T_2}$ is \textbf{blocked}, thus $F_{r_{T2}}$ shares similar facial reflection with $F_{r_{T1}}$ and can pass the $Cls$.
			$4$) The bottom row indicates the conventional spoofing case.}
		\label{ModalitySpoofing}
	\end{figure}

	\Paragraph{Light CAPTCHA Regression Branch.}\label{CAPTCHA}
	Although our system can well handle most normal $2$D or $3$D presentation attacks via disentangling depth and material information, it may still suffer from one special spoofing attack named \textit{modality spoofing}, which directly forges the desired reflection patterns.
	Specifically, modality spoofing will fail our classification net when meeting $2$ requirement:
	$1$) The formerly captured raw video consists of facial reflection frames that contains the true reflection patterns, which is leaked and replayed by Hi-res screen.
	$2$) Within the capture process of attack trial, the casted light doesn't interfere with the original facial reflection in video frames. 
	Fig.~\ref{ModalitySpoofing} illustrates the principle of our light CAPTCHA against the modality spoofing.
	We further conduct experiments to prove the effectiveness of our light CAPTCHA checking mechanism in Fig.~\ref{LightRegressionExp}.
	The $|SNR|$ results of various clients are all above $20dB$ and close with the ground truth CAPTCHA, which demonstrates its ability to distinguish $4$ types of casting light. 
	Since the fixed video loop must match the randomly generated CAPTCHA to bypass our system, this checking mechanism highly improves the security on modality spoofing.
	
	\begin{table} [t!]
		\small
		\centering
		\setlength{\abovecaptionskip}{2mm}
		\scalebox{0.92}{
			\begin{tabular}{| P{0.9cm} | P{1.46cm} | P{1.46cm} | P{1.46cm} | P{1.46cm} |}
				\hline
				\diagbox[width=4.2em, height=2em]{$\lambda_{mat}$}{$\lambda_{dep}$}    &  $0.0$     & $0.5$   & $1.0$      & $3.0$      \\ \hline
				0.0             & $1.90\pm0.25$   & $1.73\pm0.14$    & $1.80\pm0.24$ & $1.93\pm0.28$          \\ \hline
				0.5            & $1.76\pm0.15$   & $\textbf{1.37}\pm\textbf{0.10}$    & $1.48\pm0.22$ & $1.80\pm0.19$       \\ \hline
				1.0           & $2.40\pm0.25$   & $1.63\pm0.13$    & $1.60\pm0.29$ & $1.95\pm0.28$     \\ \hline
				3.0           & $2.68\pm0.47$   & $1.88\pm0.26$    & $2.21\pm0.13$ & $2.30\pm0.34$     \\ \hline
			\end{tabular}
		}
		\caption{\small Comparison of $EER$ from validation set in our dataset under different combination of hyper-parameters.}
		\label{DepthLossExp}
		\vspace{-2.3mm}
	\end{table}
	
	\Paragraph{Sensitivity Analysis.}\label{Sensitivity}
	Also, we implement a grid search on hyper-parameters to demonstrate the insensitivity of the proposed system.
	To be specific, we adjust the weight of depth supervision and material supervision in Eq.~\ref{eq:ReconLoss} and train multiple models, respectively.
	Under each $\lambda_{dep}$ and $\lambda_{mat}$, we train $10$ different models, and then evaluate the mean and standard variance of $EER$, as shown in Tab.~\ref{DepthLossExp}.
	When $\lambda_{dep}$=$0$ and $\lambda_{mat}$=$0$, the normal cues are directly used for liveness classification, which achieves the worst results.
	As we increase $\lambda_{dep}$ and  $\lambda_{mat}$ synchronously to a certain range $[0.5, 1.0]$, the performance hits the peak, which verifies its effectiveness of disentanglement to help consolidate the normal cues and enhance the representative information.
	
	\begin{figure}[t!]
		\centering
		\setlength{\lineskip}{\medskipamount}
		\setlength{\abovecaptionskip}{2mm}
		\includegraphics[width=0.9\linewidth]{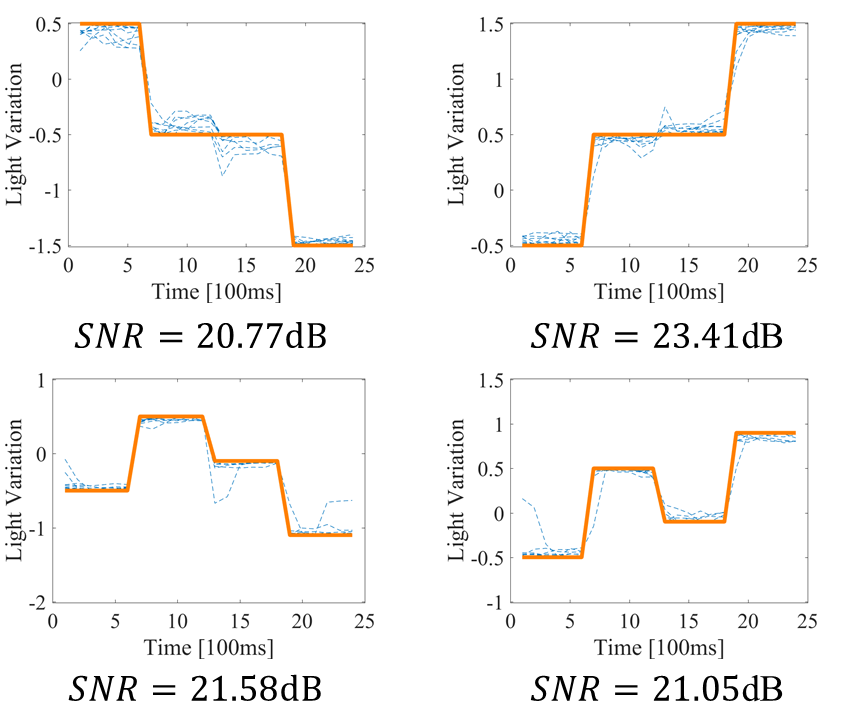}
		\vspace{-2mm}
		\caption{\small
			\textbf{Illustration on estimated light CAPTCHA}.
			Each figure shows $10$ estimated curves obtained by our regression net (blue dotted) from different subjects and scenes compared to the ground truth (orange solid), where the x-axis and y-axis denote the time and temporal variation of light hue $\alpha$ respectively. }
		\label{LightRegressionExp}
	\end{figure}
	
	\subsection{Comparison to State-of-the-Art}\label{4.4}
	
	\begin{table}[t!]
		\small
		\centering
		\setlength{\abovecaptionskip}{2mm}
		\scalebox{1}{
			\begin{tabular}{|c|c|c|c|c|}
				\hline
				Method       & EER~(\%)     & HTER~(\%)     \\ \hline
				SURF~(Boulkenafet et al.)& $4.72$     & $14.65$                                         \\ \hline
				Deep LBP~(Li et al.)& $5.61$     & $8.83$                                            \\ \hline
				FASNet~(Lucena et al.)& $5.67$       &  $8.60$                                            \\ \hline
				Noise Modeling~(Jourabloo et al.)&   $4.80$    &         $4.85$        \\ \hline
				Auxiliary Depth CNN~(Liu et al.)& $2.27$      & $2.96$                \\ \hline
				Ours      & $\textbf{1.24}$   & $\textbf{1.91}$                          \\ \hline
			\end{tabular}
		}
		\caption{\small Comparison of $EER$ from validation set and $HTER$ from test set in our dataset.}
		\label{HTER}
		\vspace{-2mm}
	\end{table}
	
	\Paragraph{Face Anti-Spoofing.}
	We conduct comparisons on anti-spoofing, in which our method and several state-of-the-art methods are trained on our dataset, and then tested on Replay-Attack, CASIA datsets and our dataset, respectively.
	After training, we determine the threshold $\tau_{cls}$ via the $EER$ on the validation set and evaluate the $HTER$ on the test set.
	First, we conduct test on our dataset.
	Tab.~\ref{HTER} shows that our method significantly outperforms the prior methods, where Aux Depth~\cite{liu2018learning} ranks $2$nd, while the conventional texture based methods~\cite{boulkenafet2017face,li2017face} achieve relatively lower performance.
	
	Next, we conduct tests on two public datasets: Replay-Attack~\cite{DBLP:conf/biosig/ChingovskaAM12} and CASIA~\cite{casia2012face}.
	To better show the generalization of our method, \emph{no} additional fine-tuning is performed.
	Since our method requires casting extra light onto the subjects, the only way to test the live subjects is to let the real person involved in the public dataset to be presented, which is impossible and unable for us to measure $FRR$ on public dataset.
	For the spoof samples in these two public datasets, we print or broadcast the videos to act as the negative subjects and evaluate the $FAR$ of various methods in Tab.~\ref{FAR on public set}.
	The results again demonstrate the effectiveness and generalization of our method compared to the state-of-the-art methods.
	
	\begin{figure}[t!]
		\centering
		\setlength{\lineskip}{\medskipamount}
		\setlength{\abovecaptionskip}{2mm}
		\setlength{\belowcaptionskip}{0mm}
		\includegraphics[width=0.95\linewidth]{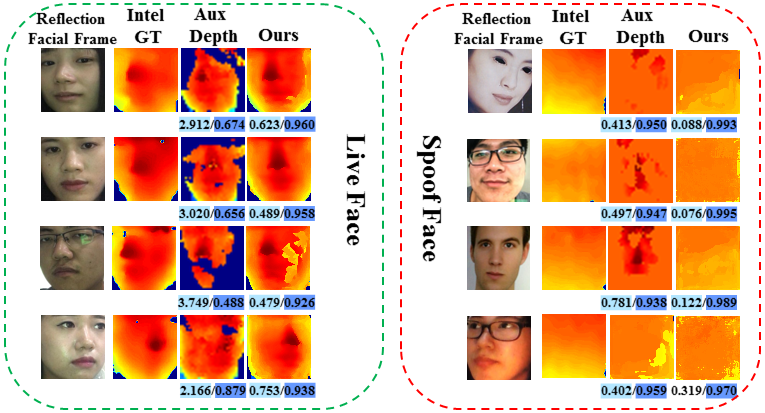}
		\caption{\small
			\textbf{Comparisons on depth recovery}.
			We take the depth data from Intel $3$D camera as the ground truth.
			Results are computed using the depth metrics from~\cite{godard2017unsupervised}.
			The light blue $RMSE(log)$ measures error in depth values from the ground truth (Lower is better).
			And the dark blue $\delta < 1.25$ measures error in the percentage of depths that are within threshold from the correct value (Higher is better).
			Note that Aux Depth~\cite{liu2018learning} recovers depth map from single RGB image, while ours is recovered from reflection frames which contain solid depth clues.
		}
		\label{DepthMapRecover}
	\end{figure}

	\begin{table} [t!]
		\small
		\centering
		\setlength{\abovecaptionskip}{2mm}
		\scalebox{0.95}{
			\begin{tabular}{|c|c|c|}
				\hline
				\multirow{2}{*}{Method}    & Replay-Attack        & CASIA                 \\ \cline{2-3}
				& FAR(\%) & FAR(\%) \\ \hline
				Color texture~(Boulkenafet et al.)& $0.40$                            & $6.20$                     \\ \hline
				Fine-tuned VGG-face~(Li et al.)& $8.40$                            & $5.20$                     \\ \hline
				DPCNN~(Li et al.)& $2.90$                            & $4.50$                      \\ \hline
				SURF~(Boulkenafet et al.)& $0.10$                             & $2.80$                      \\ \hline
				Deep LBP~(Li et al.)& $0.10$          & $2.30$                      \\ \hline
				Patch-Depth CNNs~(Atoum et al.)& $0.79$                              & $2.67$                      \\ \hline
				Ours     & $\textbf{0.02}$          & $\textbf{0.75}$    \\ \hline
			\end{tabular}
		}
		\caption{\small $FAR$ indicator cross-tested on public dataset. Here to mention we use the same model trained from our dataset without finetuning and same $\tau_{cls}$ to evaluate $FAR$ on public dataset.}
		\label{FAR on public set}
		\vspace{-2.5mm}
	\end{table}
	
	\Paragraph{Visualization.}
	We conduct comparisons on depth recovery against the recent state-of-the-art method~\cite{liu2018learning}, as shown in Fig.~\ref{DepthMapRecover}.
	Our method can recover more accurate depth map on various aspects, such as pose, facial contour and organ details, which demonstrates the effects to recover depth from solid depth clue instead of RGB texture.
	Further, our method achieves comparable results to the Intel $3$D sensor that can absolutely detect $2$D presentation attacks without failure cases.
	
\begin{figure}[t!]
		\centering
		\setlength{\lineskip}{\medskipamount}
		\setlength{\abovecaptionskip}{2mm}
		\includegraphics[width=0.99\linewidth]{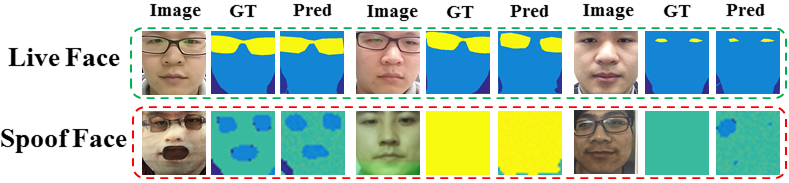}
		\vspace{-2mm}
		\caption{\small
			\textbf{Visualization on material recovery}. The top line shows three live faces, while the bottom line displays $3$D, screen and paper attack, respectively. Environment: dark blue. Real Face: blue. Paper: green. Eye/Screen: Light yellow.
		}
		\label{MaterialMapRecover}
	\end{figure}

	We further visualize the estimated material map with the pre-defined ground truth, as shown in Fig.~\ref{MaterialMapRecover}.
	To generate the ground truth material map, we construct a pixel-wise mapping from material to brightness, where the material with low albedo is mapped into low brightness and vice versa (\emph{e.g.}, environment is indicated in dark color, and screen is indicated in light color.). Under this rule, the proposed system can further generalize in unseen materials, if the albedo of unseen material is comparable with the existing four materials in our dataset. 
	
	\begin{figure}[t!]
		\centering
		\setlength{\lineskip}{\medskipamount}
		\setlength{\abovecaptionskip}{2mm}
		\setlength{\belowcaptionskip}{-2mm}
		\includegraphics[width=0.88\linewidth]{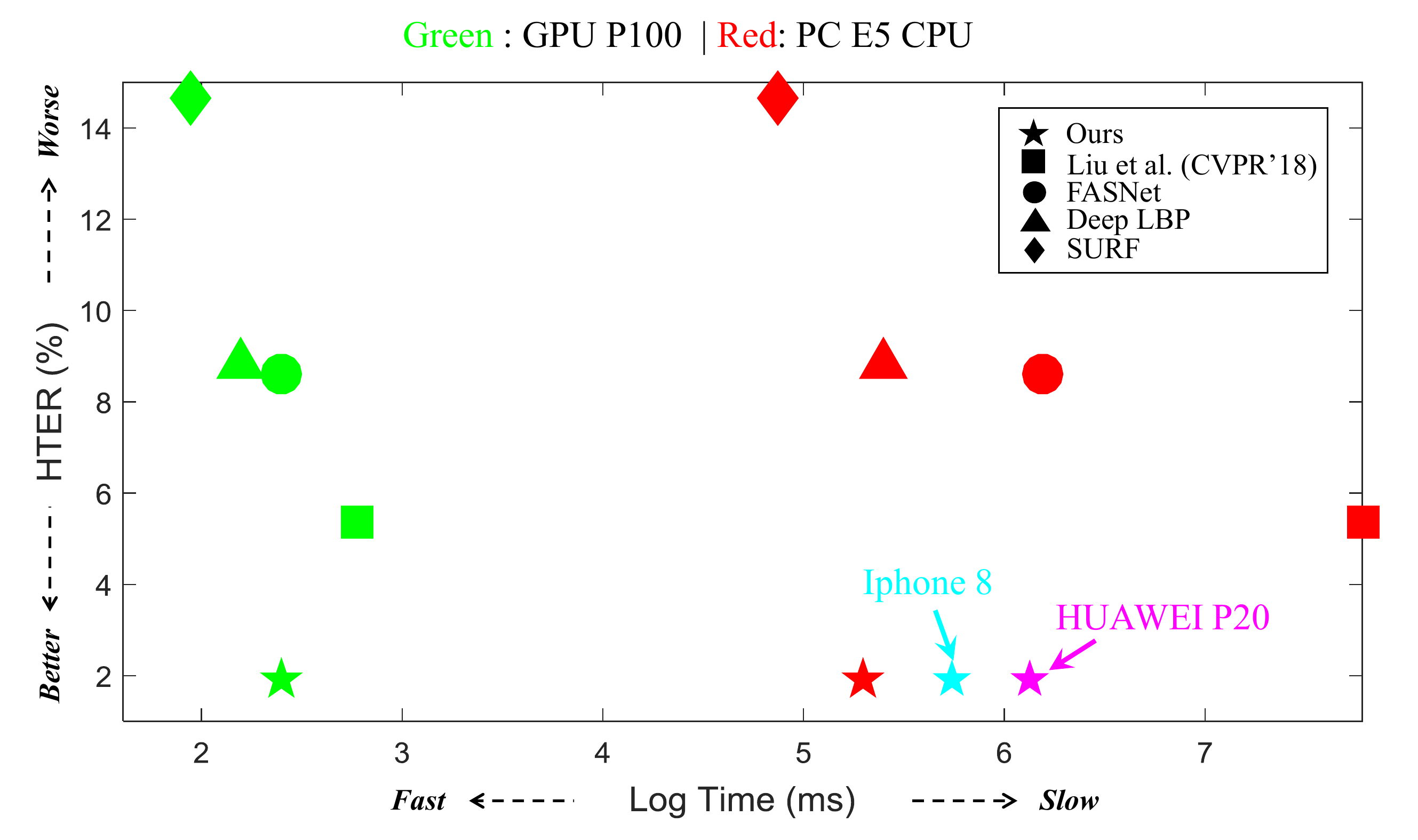}
		\caption{\small\textbf{ Time Comparison} between several SOTA methods and ours in the aspects of effectiveness and cross-platform efficiency.}
		\label{TimeComplexity}
	\end{figure}
	
	\Paragraph{Running Time Comparison.}
	We compare the cross-platform inference time with several state-of-the-art methods.
	We deploy and compare on $3$ common platform architectures: GPU for cloud server, CPU (x$86$) for some embedded chips and CPU (arm) for smart phones. As shown in Fig.~\ref{TimeComplexity},
	our efficiency on mobile platform still meets the application requirement, and even outperforms some methods on CPU (x$86$).
	The results indicate that our method achieves real-time efficiency and is portable for cross-platform computation requirements with state-of-the-art anti-spoofing performance.

	\section{Conclusion}
	In this paper, an effective facial anti-spoofing method named Aurora Guard is proposed, which holds real-time cross-platform applicability.
	The key novelty of our method is to leverage two kinds of auxiliary information, the depth map and the material map, which are proven to be extracted from reflection frames and thus significantly improve the accuracy of anti-spoofing system against unlimited presentation attacks. Meanwhile, the light CAPTCHA checking mechanism strengthens reliability of the final judgement considering the modality spoofing.
	Extensive experiments on public benchmark and our dataset show that AG is superior to the state of the art methods.
	
	\newpage
	
	\bibliographystyle{aaai}
	\bibliography{main}
	
\end{document}